\documentclass[runningheads]{llncs}
\usepackage{graphicx}

\usepackage{tikz}
\usepackage{comment}
\usepackage{color}
\usepackage{multirow}
\usepackage{epsfig}
\usepackage{indentfirst}
\usepackage{subcaption}
\usepackage{floatrow}

\usepackage[accsupp]{axessibility}  

\begin{document}
\pagestyle{headings}
\mainmatter
\def\ECCVSubNumber{3719}  

\title{PartImageNet: A Large, High-Quality Dataset of Parts} 

\titlerunning{PartImageNet}
%
\author{Ju He\inst{1} \and
Shuo Yang\inst{2} \and
Shaokang Yang\inst{3} \and
Adam Kortylewski\inst{1,4,5} \and
Xiaoding Yuan\inst{1} \and
Jie-Neng Chen\inst{1} \and
Shuai Liu\inst{3} \and
Cheng Yang\inst{3} \and
Qihang Yu\inst{1} \and
Alan Yuille\inst{1}}
\authorrunning{J. He et al.}
%
\institute{$^1$Johns Hopkins University
$^2$University of Technology Sydney
$^3$ByteDance Inc.\\
$^4$Max Planck Institute for Informatics
$^5$University of Freiburg}
\maketitle

\begin{abstract}
    It is natural to represent objects in terms of their parts. This has the potential to improve the performance of algorithms for object recognition and segmentation but can also help for downstream tasks like activity recognition. Research on part-based models, however, is hindered by the lack of datasets with per-pixel part annotations. This is partly due to the difficulty and high cost of annotating object parts so it has rarely been done except for humans (where there exists a big literature on part-based models). To help address this problem, we propose PartImageNet, a large, high-quality dataset with part segmentation annotations. It consists of $158$ classes from ImageNet with approximately $24,000$ images. PartImageNet is unique because it offers part-level annotations on a general set of classes including non-rigid, articulated objects, while having an order of magnitude larger size compared to existing part datasets (excluding datasets of humans). It can be utilized for many vision tasks including Object Segmentation, Semantic Part Segmentation, Few-shot Learning and Part Discovery. We conduct comprehensive experiments which study these tasks and set up a set of baselines.
\end{abstract}
\section{Introduction}

\begin{figure*}
    \centering
    \includegraphics[width=\linewidth,height=5cm]{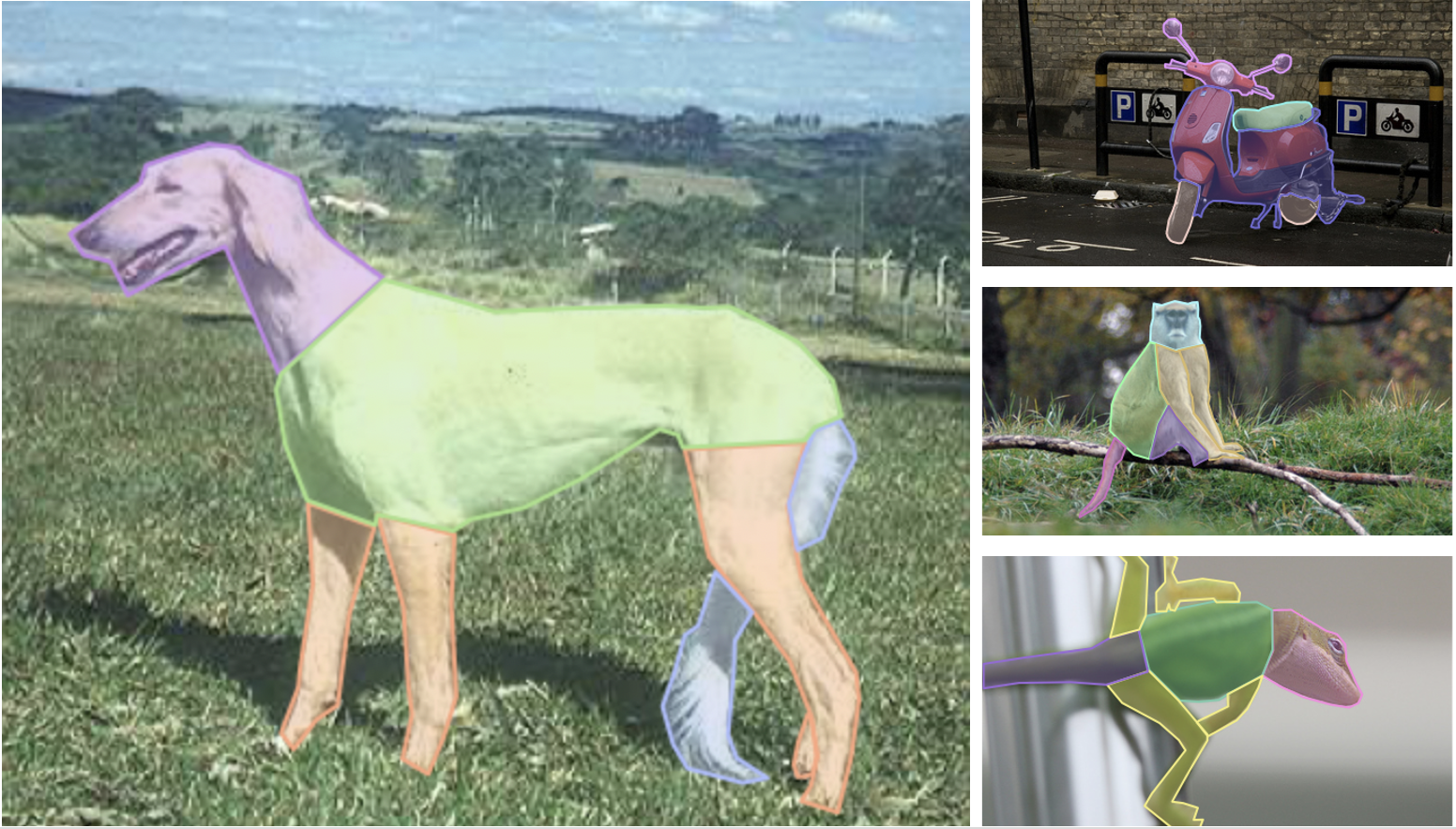}
    \caption{Example figures of annotated images in PartImageNet. We offer high-quality precise and dense part segmentation annotation on a wide range of general species including both non-rigid and rigid objects. In total there are around $24,000$ images of 158 classes from ImageNet annotated.}
    \label{fig:intro}
\end{figure*}

\indent When humans observe objects we can effortlessly parse them into their component parts. Studies in cognitive psychology show that humans learn rich hierarchical representations of objects \cite{doi:10.1126/science.aab3050} and can decompose objects into parts taking into account their spatial relationships \cite{biederman1987recognition}. Partly inspired by these findings,  computer vision researchers have studied  how to model parts  and to represent objects in terms of parts. The big literature on these topics includes deformable templates \cite{yuille1992feature}, pictorial structures \cite{felzenszwalb2005pictorial}, constellation models \cite{weber2000unsupervised,fei2006one} and grammar-based models \cite{zhu2007stochastic,girshick2011object}. In particular, there have been work \cite{xia2017joint,wang2015joint} on segmenting object parts and studying whether this helps improve the segmentation of objects, and later studies of the use of these part segmentation for downstream tasks, e.g.,  describing pedestrians and facilitating person retrieval \cite{sun2018beyond}. Most recently, part representations have been proposed to improve panoptic segmentation \cite{de2021part}. It has also been argued that object-part models will play an important role in few-shot learning \cite{he2021compas,wu2021task,xu2020attentional} and hence help to alleviate the the computer vision communities dependence on annotated large-scale datasets. In short, object-part models are promising for improving different tasks.



But in the big data era, research on part-based models and their applications is  hindered due to the shortage of datasets with per-pixel part annotations. Current datasets are almost always restricted to humans or to a small number of object categories, e.g., PASCAL-Part \cite{chen2014detect}. There is a need to extend this datasets to include many more object categories, as suggested by \cite{de2021part}. Existing part datasets almost always focus on humans or a few rigid classes such as cars. This is partly caused by the  difficulty of per-pixel part annotations compared to other types of annotation like bounding boxes. It requires much more effort to ensure accuracy and quality consistency, especially for non-rigid objects. The few works \cite{sun2018beyond,xu2020attentional,he2021transfg} that attempt to use parts as a mid-level representation, often learnt without supervision, also suffer from this lack of annotated datasets which makes it hard to evaluate whether they have actually captured meaningful object parts.

This motivates us to introduce PartImageNet — a large, high-quality dataset with part annotation on a general set of object classes. Concretely speaking, we manually select 158 classes from ImageNet \cite{deng2009imagenet} and group them into 11 super-category following the WordNet hierarchy of ImageNet. Part labels are designed according to the super-category while can be elaborated to fine-grained classes in a hierarchical way. A carefully designed pipeline is taken to ensure the high quality of our PartImageNet annotations. As far as we know, this is the only dataset after PASCAL-Part \cite{chen2014detect} that offers part-level annotations on more general classes instead of just humans and rigid objects. Compared to PASCAL-Part \cite{chen2014detect}, we annotate much more images (24k v.s. 10k) on much more classes (158 v.s. 20). Extensive experiments on PartImageNet are conducted to show that parts could help general object segmentation and few-shot learning and set up a set of baselines of different downstream tasks on this benchmark. We believe that with this dataset, the research on part-based models and their applications will be facilitated a lot. In summary, we make the following contributions in this work:
\begin{enumerate}
    \item We briefly review the history of part-based models and introduce their potential applications in downstream tasks.
    \item We introduce PartImageNet — a large, high-quality dataset with part annotations on a general set of classes. From our perspective, part-level annotation, especially on non-rigid objects, is very rare and valuable.
    \item We set up a set of baselines on PartImageNet in different vision tasks with state-of-the-art methods which shows the broad usage of the dataset.
    \item We conduct experiments to show that introducing parts annotations is beneficial to the object segmentation and few-shot learning which points out to a promising future direction.
\end{enumerate}
\section{Related Work}

\subsection{Part-based Models} 

Modeling objects in terms of parts is a long-standing problem in computer vision and there is rich history of research on this topic. Starting from Pictorial Structures in the early 1970's \cite{fischler1973representation}, plenty of different models \cite{yuille1992feature,felzenszwalb2005pictorial,weber2000unsupervised,fei2006one,zhu2007stochastic,girshick2011object} have been proposed to explicitly model parts and their spatial relations to the whole object. There have been a variety of technical approaches but a common theme is that object-part models  provide rich representations of objects and help interpretablity. In recent years, partly due to the availability of big data, research on part-based models also includes part segmentation and unsupervised part exploitation. 


\noindent \textbf{Supervised Part Segmentation} 
Annotated human parts datasets have long existed and there has been much work \cite{chen2014articulated,xia2017joint} on human part detection and semantic segmentation of human parts. Sun et al. \cite{sun2018beyond} proposed to exploit part-level features for pedestrian image description and thus facilitate person retrieval. There has also been some work \cite{xia2017joint,wang2015joint} on semantic segmentation of parts of a limited class of other objects.


\noindent \textbf{Unsupervised Part Exploitation on general tasks}
Due to the lack of annotated data on a more general set of classes, research on parts for non-human object classes is often unsupervised. Thewlis et al. \cite{thewlis2017unsupervised} proposed to enforce the equivariance of landmark locations under artificial transformations of images to generate semantic meaningful parts of objects. Lorenz et al. \cite{Lorenz_2019_CVPR}  improved part discovery by simultaneously exploiting the invariance and equivariance constraints between synthetically transformed images and disentangling the shape and appearance of objects.
Recently, parts have shown to be beneficial to unsupervised, or few-shot, object learning since the modeling of parts helps alleviates the scarcity of training data provided the parts can be shared between different  classes. He et al. \cite{he2021compas} exploited the fact that the feature vectors in the CNN can be viewed approximately as part detectors and their geometry relations could be estimated by clustering. Additionally, it has been shown that few-shot segmentation benefits from the modeling of parts. Liu et al. \cite{liu2020part} decomposes the holistic class representation into a set of part-aware prototypes, capable of capturing diverse and fine-grained object features, which benefits semantic segmentation.

However, the part modeling of these methods mainly relies on unsupervised clustering combined with attention mechanisms. Without strong supervision, the results produced by these methods are not very satisfactory, because they only generate meaningful part segmentation in a few simple scenarios. It remains unclear if these approaches can really lead to the learning of semantically meaningful parts without evaluation. Thus it is important to introduce dataset with part annotation to analyze the actual effectiveness of such part modeling and promote the further research on this promising direction.

\subsection{Part Datasets}

\noindent \textbf{2D Part Datasets}
There exists multiple ways to annotate object parts in images. Among them, bounding boxes and keypoints are relatively easy to annotate while per-pixel segmentation is much harder due to the extreme fine-grained difficulty and high cost. The type of objects (i.e. rigid v.s. non-rigid) also plays an important role in deciding the difficulty of annotation. Wang et al. \cite{wang2015unsupervised} provides dense bounding box annotation for parts on 6 vehicle categories. PASCAL3D+ \cite{xiang2014beyond}, CarFusion \cite{reddy2018carfusion} and Apollocar3D \cite{song2019apollocar3d} offer keypoint annotation on a few rigid object categories especially on cars. ADE20K \cite{zhou2017scene} contains part segmentation annotation on many rigid object categories. As for non-rigid objects, CUB-200-2011 \cite{WahCUB_200_2011} provides keypoint location for birds parts. LIP \cite{liang2018look}, MHP \cite{zhao2018understanding}, CIHP \cite{gong2018instance} include instance-aware, part-level annotations for human. 

As far as we are concerned, PASCAL-Part \cite{chen2014detect} is the only existing dataset that offers part-level annotation on a more general set of categories in a per-pixel segmentation format. However, it contains relatively small number of images and only a small set of classes. We introduce a much larger,  high-quality dataset of parts to support more research on part-based models.

\noindent \textbf{3D Part Dataset}
3D Part datasets also play a crucial role in the advances of 3D shape understanding tasks. Yi et al. \cite{yi2016scalable} first takes an active learning approach to annotate the 3D models selected from 16 categories in ShapeNet \cite{chang2015shapenet}. PartNet \cite{mo2019partnet} further provides hierarchical part annotations on 3D models covering 24 object categories, most of which are indoor furniture. Recently, CGPart \cite{liu2021cgpart} proposes to use computer graphics model to efficiently generate a large-scale vehicle dataset which offers part annotation. Although these datasets have shown their effectiveness in helping data-driven models, they still suffer from the problem of a lacking annotation of non-rigid object categories. The different domains also limit their usage in the image part-level parsing.
\section{PartImageNet Dataset}

\begin{figure*}
    \centering
    \includegraphics[width=\linewidth]{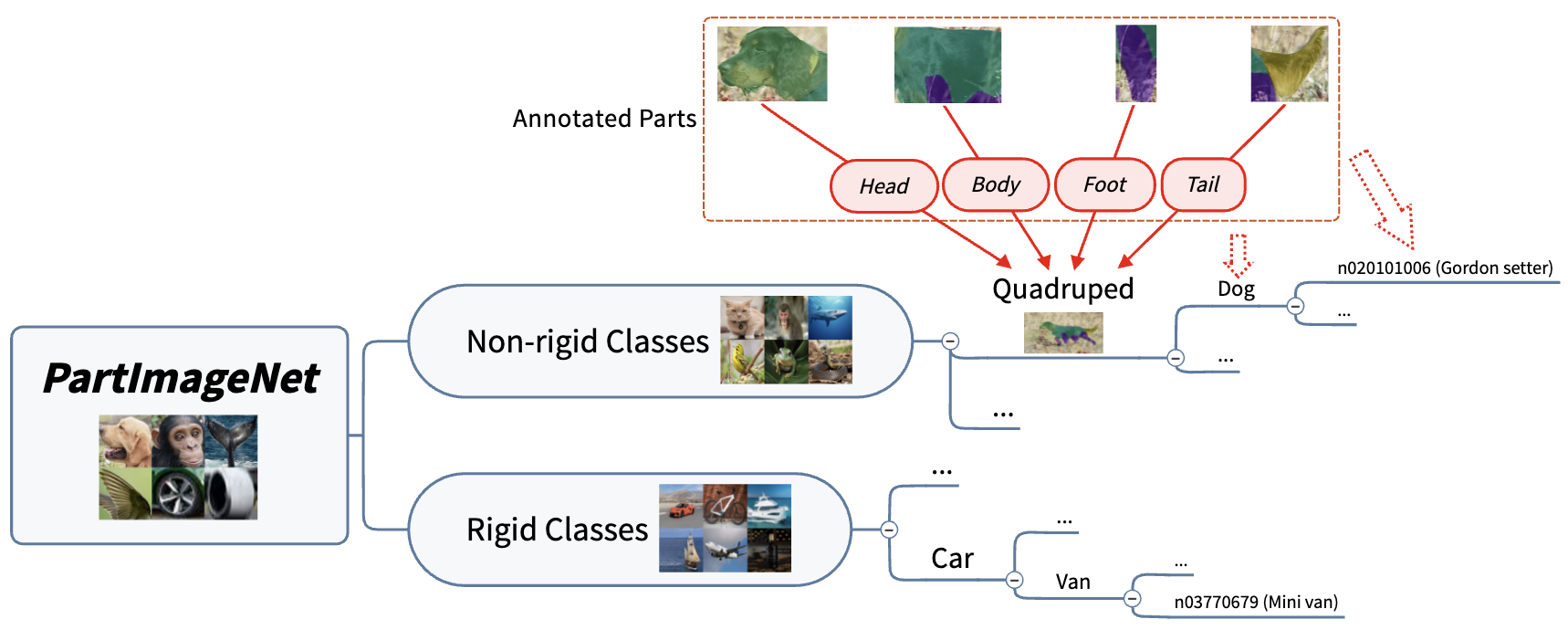}
    \caption{Overview of the PartImageNet dataset. Accurate part segmentation masks on both non-rigid and rigid objects are offered. The part labels are defined on the super-category (e.g. Quadruped) level but can be easily transferred to mid-level or class-level part labels according to the need (e.g Quadruped Head $\rightarrow$ Dog Head $\rightarrow$ Gordon setter Head) following the WordNet hierarchy as shown by the red dotted arrow.}
    \label{fig:dataset}
\end{figure*}

In this section, we present the details of how we collect and annotate the data, followed by statistics and analyze on the quality of the PartImageNet dataset. The overview of the PartImageNet is shown in Figure \ref{fig:dataset}.

\subsection{Data Collection}
\subsubsection{Data Source}
As suggested by the name, we collect images from the ILSVRC-12 dataset \cite{deng2009imagenet}. There are in total 158 object classes selected to be annotated in our dataset. All the images conform to licensing for research purposes. 

\subsubsection{Object Categories} 
\label{sec:category}
Analogous to tieredImageNet \cite{ren2018meta}, which is also a subset of ILSVRC-12 \cite{deng2009imagenet}, we group classes into super-categories corresponding to higher-level nodes in the ImageNet hierarchy. There are 11 super-categories in total. To make the dataset more challenging and more valuable, we pick up fewer rigid objects such as vehicle but choose more animals to annotate. Thus for super-categories like Quadruped, they contain around 40 classes while for super-categories such as vehicle, boat, they only have less than 10 classes. In total, there exists 118 classes out of 158 which are non-rigid objects. 

\subsubsection{Filtering Unsuitable Images}
\label{sec:filter}

As our PartImageNet dataset focuses on part segmentation, we eliminate those images which have no proper parts to annotate due to possible occlusion or improper viewpoints. Besides, to simplify the annotation process and avoid ambiguity during annotation, we discard all the images that contain more than one desired object with qualified parts to be annotated. (i.e. all annotated parts in an image are guaranteed to belong to one object). 

\subsection{Annotation}
Instead of annotating bounding boxes or keypoints which is much easier, we aim at annotating high-quality part segmentation mask at pixel-level. The part labels for classes are determined by the corresponding super-categories (i.e. classes under the super-category share the same part labels). It is difficult to ensure that all the classes under the same semantic category have the same detailed parts so we only annotate those that are most important. Take horns for example, some mammals have horns while others do not, to simplify the definition and the annotation process, we do not create a horn label for mammals. Instead, only head label exists and for mammals with horns, the horns are also counted as part of the head during annotation. The detailed part labels for different categories are shown in Table \ref{tab:annotated_parts}. Notice that though the part labels are defined at the super-category level, they can easily be transferred to mid-level or class-level part labels according to the need following the WordNet hierarchy of the ImageNet as shown in Figure \ref{fig:dataset}. 

\subsubsection{Annotation Pipeline}
\label{sec:pipeline}
Due to the extreme difficulty of annotating part segmentation mask at pixel level, a good annotation pipeline is of great importance to ensure the high quality and maintain the consistency of the annotation. Motivated by existing works on datasets such as ImageNet \cite{deng2009imagenet}, COCO \cite{lin2014microsoft}, Objects365 \cite{shao2019objects365}, we divide our annotation pipeline into three steps. First as we sample our images from ImageNet, we already have the class and super-category label information for the images, thus we split the annotation task into 11 (equals to the number of super-categories) sub-tasks to alleviate the workload of annotators. The second step is to choose whether to keep the image for the PartImageNet or not following the requirements in Sec \ref{sec:filter}. In the last step, the annotators are going to annotate the segmentation mask of specific parts for the corresponding super-category. The specific part labels to be annotated are automatically generated according to the super-category label, thus the annotators do not need to pick the right parts among all possible parts of other super-categories which significantly improves the overall accuracy and efficiency.

\subsubsection{Annotation Team}
We divide our annotation team into three groups: annotators, inspectors, examiners. All the images are first annotated by annotators, then a subset of annotated images will be randomly chosen to be checked by the inspectors. In the end, the examiners can check as many images as they want and see if the quality meets the requirement. Any failures in the above steps will result in re-annotation of the job. 

\noindent \textbf{Annotator} The annotators' job is to  annotate all the images following the pipeline introduced in Sec \ref{sec:pipeline}. Before starting the official annotation, all annotators are going to take courses from the inspectors and go for a test annotation round. During the whole annotation process, the annotators can contact the inspectors at any time if they have questions regarding the current task. 

\noindent \textbf{Inspector} The duty of the inspectors is to ensure the quality of the annotation made by the annotators. The inspectors will first have a meeting with the annotators before annotation to teach them the annotation rules. Then they will review all the annotated images during the test annotation round and provide feedback to the annotators. For the official annotation, a random subset of annotated images will also be reviewed by them. If there is an obvious error or the annotation fails to meet the quality requirement, the task will be rejected and re-annotated. If the rejection ratio of an annotator is too high, then all his annotated tasks will be discarded and assigned to other annotators.

\noindent \textbf{Examiner} Examiners design the annotation rules and discuss with the inspectors to make the quality requirement. They are also responsible for answering all the ambiguity questions. After all annotations are done, they review most of the annotated images to ensure the quality and unqualified ones will be rejected and re-annotated.


\subsection{Annotation Quality and  Consistency}
Part segmentation annotation is very hard due to the variance of objects pose, orientation, occlusion. Besides, another important problem that does not exist in other annotation tasks is the ambiguity of how to define the separation of different parts (i.e. how do we define the boundary between head and body). It is impossible to make clear annotation rules that can keep all annotators consistent on the boundary annotation and handle so many different variations. Thus, to make the annotation of parts as accurate and consistent as we can, we make the following efforts:

\noindent \textbf{Maximize possible information} The annotators are asked to annotate all visible and distinguishable parts no matter how small they are in order to keep as much information as we can. We believe that such kind of small parts also need to be studied and should be handled during the algorithm design stage instead of the annotation stage.

\noindent \textbf{Accurate boundary segmentation} To keep the annotations accurate so they do not contain too many background pixels. We set strict annotation rules to guide the annotators to create tight segmentation masks in most situations (the requirements are relaxed under fuzzy situations). The occluders are also required to be masked out of the annotation.  

\noindent \textbf{Consistent Annotation Task Assignment} As our part labels are set according to the super-categories of the images. To keep the consistency among images of the same super-category, we divide the annotators into sub-groups where each sub-group is responsible for annotating one super-category and can communicate freely within the group. In this way, we alleviate the inconsistency of boundary separation especially among the same super-category.

\subsection{Statistics}

\subsubsection{Category and Class Mapping}

\begin{table*}[]
\small
\centering
\caption{Number of classes and annotated parts for each category in PartImageNet. The number in the brackets after the category name indicates the total number of classes under the category.}
\label{tab:annotated_parts}
\begin{tabular}{c|c}
\hline
Category & Annotated Parts \\ \hline
Quadruped (46) & Head, Body, Foot, Tail \\
Biped (17) & Head, Body, Hand, Foot, Tail \\
Fish (10) & Head, Body, Fin, Tail \\
Bird (14) & Head, Body, Wing, Foot, Tail \\
Snake (15) & Head, Body \\
Reptile (20) & Head, Body, Foot, Tail \\
Car (23) & Body, Tire, Side Mirror \\
Bicycle (6) & Head, Body, Seat, Tire \\
Boat (4) & Body, Sail \\
Aeroplane (2) & Head, Body, Wing, Engine, Tail \\
Bottle (5) & Body, Mouth \\ \hline
\end{tabular}
\end{table*}

As introduced in Sec \ref{sec:category}, our PartImageNet focuses more on the challenging animals categories instead of rigid object categories. Thus we pick far more classes in animals than cars, boats, planes etc. The detailed number of classes per category is shown in Tab \ref{tab:annotated_parts}. In summary, we annotated 158 classes and 118 of them are non-rigid objects. By contrast, PASCAL-Part \cite{chen2014detect} annotated 20 classes and 12 of them are non-rigid objects. 

\subsubsection{Total Annotated Images}


Based on our proposed annotation pipeline, around 24,000 images are annotated in the PartImageNet dataset. We randomly sampled $85\%$, $5\%$ and $10\%$ images per class into training, validation and testing set. The detailed numbers of images and annotated parts in each set is shown in Table \ref{tab:anno_num}.

\begin{center}
\begin{minipage}{\textwidth}
    \begin{minipage}[t]{0.45\textwidth}
        \centering
        \makeatletter\def\@captype{table}\makeatother\caption{The annotation split of the PartImageNet dataset.}
            \begin{tabular}{c|cc}
                \hline
                 & Images & Parts \\ \hline
                Train & 20481 & 95059 \\
                Validation & 1206 & 5626 \\
                Test & 2408 & 11275 \\
                All & 24095 & 111960 \\ \hline
            \end{tabular}
            \label{tab:anno_num}
    \end{minipage}
    \begin{minipage}[t]{0.45\textwidth}
        \centering
        \makeatletter\def\@captype{table}\makeatother\caption{The annotation density of the PartImageNet dataset.}
            \begin{tabular}{c|c}
                \hline
                Number of parts & Proportion (\%) \\ \hline
                1-2 & 22.00 \\
                3-6 & 57.65 \\
                7-9 & 18.61 \\
                10+ & 1.74 \\ \hline
            \end{tabular}
            \label{tab:anno_density}
    \end{minipage}
\end{minipage}
\end{center}

\subsubsection{Annotation Density}

Though we directly offer more high-level part annotation compared to PASCAL-Part \cite{chen2014detect}, PartImageNet still has a high density of part annotations with 111960 part instances in total. We compute the proportion of number of annotations per image as shown in Table \ref{tab:anno_density}. Most images contain around 3-6 annotations which is quite dense and considered our overall dataset size, it should provide enough training examples for most algorithms. 

\subsubsection{Class Distribution}

\begin{figure}
    \centering
    \includegraphics[width=\linewidth]{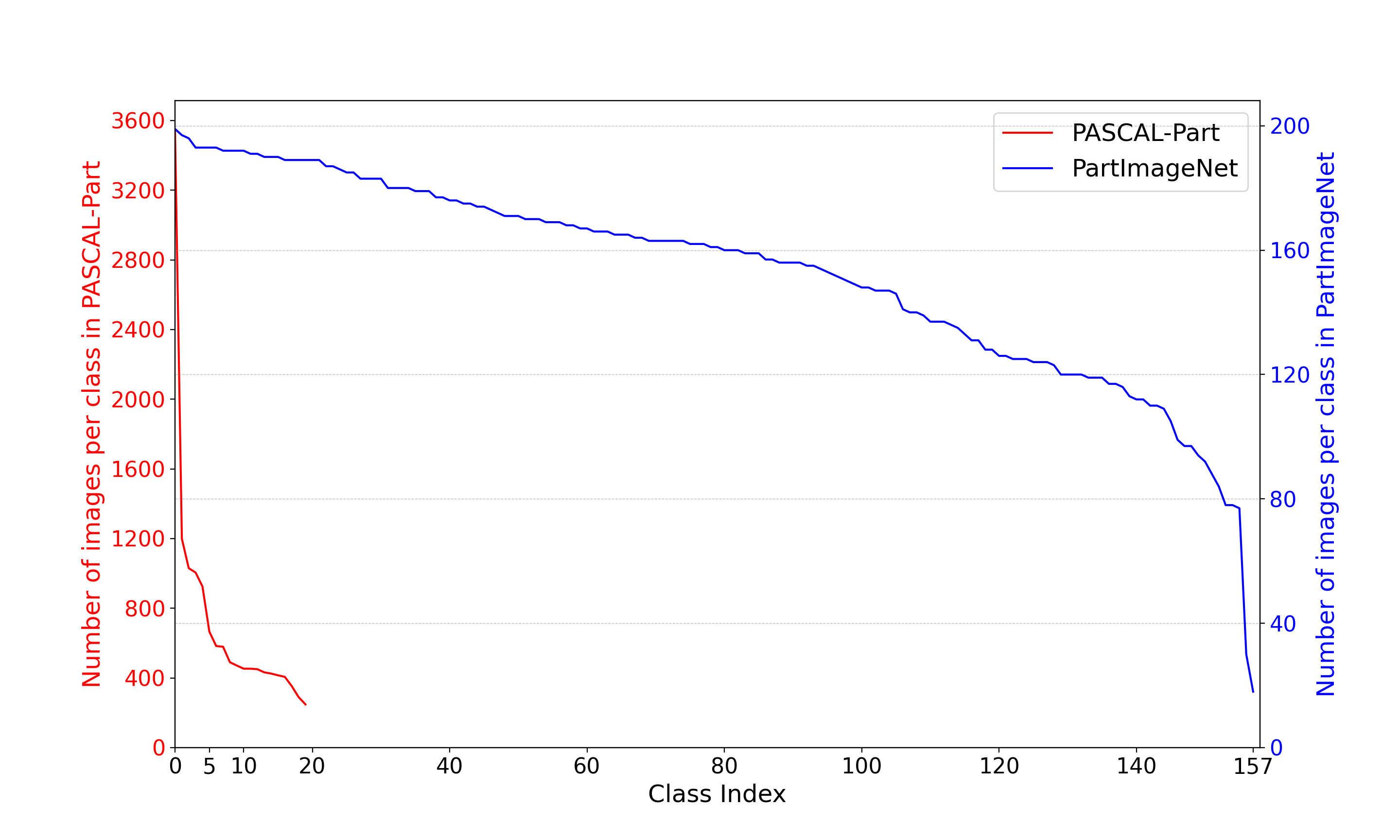}
    \caption{Number of images per class in PartImageNet and PASCAL-Part \cite{chen2014detect}. Class index is sorted according to the number of images in the class. Our PartImageNet exhibits a more balanced distribution with a few tail classes that contain few images. By contrast, PASCAL-Part exhibits a more sharpen drop tendency.}
    \label{fig:class_distribution}
\end{figure}

To dive into the details of the PartImageNet, we provide the overall class distribution in Fig \ref{fig:class_distribution}. Though the original number of images per class in ImageNet \cite{deng2009imagenet} is roughly the same. After our annotation process, some classes will have more images be ruled out due to multiple objects in one image or occlusion (e.g. bottle). Thus PartImageNet naturally has a few tail classes that contain few images. However, compared to the PASCAL-Part \cite{chen2014detect} dataset, PartImageNet exhibits a much more balanced distribution. The number of images of the most majority class in PASCAL-Part is roughly 9 times more than that of the middle class, while that ratio in PartImageNet is only 1.25.

\section{Experiments}

In this section, we conduct extensive experiments on our proposed PartImageNet for different tasks including semantic part segmentation, object segmentation and few-shot learning. We broadly evaluate classic methods along with some of the state-of-the-art models to set up a set of baselines on this benchmark. While all these methods do not take part annotations into account, we further show that by exploiting annotated parts, the performance on object segmentation and few-shot learning could get a non-trivial improvement.

\subsection{Semantic Part Segmentation}

We conduct experiments on semantic part segmentation PartImageNet using Semantic FPN \cite{ak2019panopticfpn}, Deeplabv3+ \cite{chen2018encoder} and SegFormer\cite{xie2021segformer}. Semantic FPN \cite{ak2019panopticfpn} and Deeplabv3+ \cite{chen2018encoder} are classic convolution-based semantic segmentation methods. SegFormer\cite{xie2021segformer} is one of the state-of-the-art transformer-based semantic segmentation frameworks. We use Resnet-50 as backbone for Semantic FPN\cite{ak2019panopticfpn} and Deeplabv3\cite{chen2017deeplabv3}. MiT-b2 (Mix Transformer encoders) is adopted for Segformer \cite{xie2021segformer}.

\begin{figure*}[!h]
    \centering
    \includegraphics[width=\linewidth,height=10cm]{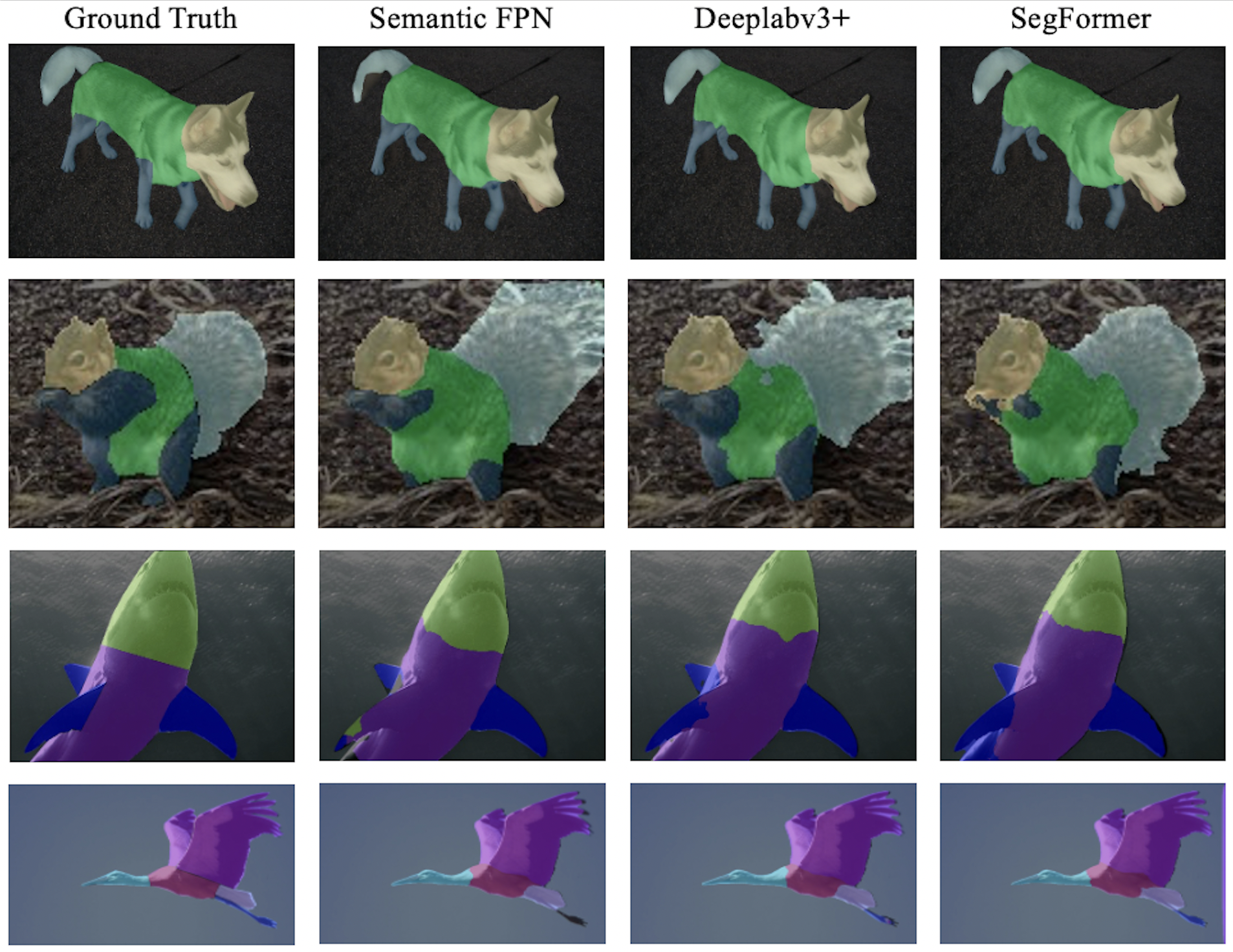}
    \caption{Example figures of semantic part segmentation results. The quality of the results highly depend on the background, shape of the objects, occlusion situation and the class itself. More visualizations are shown in supplementary materials.}
    \label{fig:vis}
\end{figure*}

Table \ref{tab:part_seg} summaries the results for semantic part segmentation and Figure \ref{fig:vis} shows the qualitatively visualizations. As can be observed, methods with strong supervision can generally produce satisfactory results on the semantic part segmentation especially when the background and the shape are relatively easy. However, they still suffer from three main challenges: 1) inaccurate boundary between semantic parts, 2) wrong label assignments on similar semantic parts, 3) ignoring small semantic parts (e.g. See row3 - row5 of Fig \ref{fig:vis}). Besides, we also notice that the recent progress in methods for general object segmentation does not seem to bring expected improvement in the context of semantic part segmentation (e.g. SegFormer only bring limited improvement based on Deeplabv3+). This reveals the difficulty of part segmentation and indicates that special architecture or module design is needed for effectively solving the challenge.

\begin{table*}[]
\small
\centering
\caption{Experimental results of Part Segmentation on PartImageNet. Performance are evaluated in terms of mIoU.}
\label{tab:part_seg}
\begin{tabular}{ccccc}
\hline
Model & Backbone & Crop Size & Val mIoU & Test mIoU \\ \hline
Semantic FPN \cite{ak2019panopticfpn}  & ResNet-50 & 512x512 & $56.76$ & $54.57$ \\
Deeplab v3+ \cite{chen2018encoder} & ResNet-50 & 512x512 & $60.57$ & $58.71$ \\
SegFormer \cite{xie2021segformer} & MiT-B2 & 512x512 & $61.97$ & $61.46$ \\ \hline
\end{tabular}
\end{table*}

\subsection{Object Segmentation}

While we offer part-level segmentation masks, they could easily be integrated to offer a full object segmentation mask and thus serves as a benchmark for Salient Object Segmentation. As this role, PartImageNet is a relatively easy one compared to popular MS-COCO \cite{lin2014microsoft} and CityScapes \cite{cordts2016cityscapes} since it is a salient holistic segmentation of a single object. However, it is unique as it offers the opportunity to conduct research on the relation between parts and whole objects as shown below. Besides, it also has the advantage that it could be evaluated at a hierarchy level (i.e. super-category \& fine-grained class).

\subsubsection{Baseline}

We still adopt Semantic FPN \cite{ak2019panopticfpn}, Deeplabv3+ \cite{chen2018encoder} and SegFormer \cite{xie2021segformer} for object segmentation. The experimental setup and training pipeline are in line with those in semantic part segmentation. 

Table \ref{tab:object_seg} summarizes the results of Object Segmentation on PartImageNet. Here the mIoU on the fine-grained classes is reported as it is more challenging than segmenting the super-category object labels. As can be observed, existing methods already achieve quite good results on the benchmark as expected. We are more interested in whether parts can help the general object segmentation.

\begin{table*}[]
\small
\centering
\caption{Experimental results of Object Segmentation on PartImageNet. Performance are evaluated in terms of mIoU.}
\label{tab:object_seg}
\begin{tabular}{ccccc}
\hline
Model & Backbone & Crop Size & Val mIoU & Test mIoU \\ \hline
Semantic FPN \cite{ak2019panopticfpn}  & ResNet-50 & 512x512 & $61.47$ & $60.08$ \\
Deeplab v3+ \cite{chen2018encoder} & ResNet-50 & 512x512 & $68.38$ & $63.99$ \\
SegFormer \cite{xie2021segformer} & MiT-B2 & 512x512 & $74.55$ & $71.13$ \\ \hline
\end{tabular}
\end{table*}
\vspace{-5mm}

\subsubsection{Can Parts help Object Segmentation?}

Motivated by the intuition that it would be a natural way for models to first learn to group similar pixels together at the early stage followed by gradually forming a whole object mask, we design experiments to validate whether object segmentation can be improved by introducing part annotations as deep supervision. We take Deeplabv3+ \cite{chen2018encoder} here for concrete analysis and ablation study.

Table \ref{tab:ds_stage} summaries the experimental results of adding part annotations as deep supervision in Deeplabv3+ \cite{chen2018encoder} at different stages of the backbone. We observe that adding part annotations at early stage such as stage 2 does not have an obvious effect on the results probably because the features are still too local without the ability to capture semantic meanings. While starting from stage 3, the deep supervision gradually increases the performance of object segmentation. When adding it at the stage 4 , the model achieves largest improvement by $1.63\%$ and $1.05\%$ mIoU on the validation and testing set respectively.

Table \ref{tab:ds_type} further conducts ablation study on the type of deep supervision when adding at the 4th stage. We first show that by adding object masks as deep supervision, the performance decreases a little which reveals that by simply introducing object segmentation mask as deep supervision, the performance can not be improved. Then we show that when adding part annotations, it should be supervised with binary cross entropy loss (i.e. only the current part class is considered) in the sense that we should encourage pixels that belong to the same part to be similar while avoiding penalizing wrong classification of the part classes. The reasons for that are two-folded: firstly semantic part segmentation is a harder task compared to object segmentation, we should not do the full part of it at the early stage of the network, secondly some pixels of different parts are very similar at the pixel level which makes them hard to be distinguished when semantic meanings have not be well-learned at the shallow stage. 

\begin{center}
\begin{minipage}{\textwidth}
    \begin{minipage}[t]{0.48\textwidth}
        \centering
        \makeatletter\def\@captype{table}\makeatother\caption{Experimental results of exploiting part segmentation as deep supervision at different stages of the backbone. DS stands for Deep Supervision.}
            \begin{tabular}{ccc}
                \hline
                DS Stage & Val mIoU & Test mIoU \\ \hline
                None & $68.38$ & $63.99$ \\
                Stage 2 & $68.98$ & $64.53$ \\
                Stage 3 & $68.84$ & $64.72$ \\
                Stage 4 & $\mathbf{69.82}$ & $\mathbf{65.59}$ \\ \hline
            \end{tabular}
            \label{tab:ds_stage}
    \end{minipage}
    \begin{minipage}[t]{0.48\textwidth}
        \centering
        \makeatletter\def\@captype{table}\makeatother\caption{Exploiting different kinds of part segmentation as deep supervision at Stage 4. CE stands for Cross Entropy and BCE stands for Binary Cross Entropy.}
            \begin{tabular}{ccc}
                \hline
                DS Type & Val mIoU & Test mIoU \\ \hline
                None & $68.38$ & $63.99$ \\
                Object & $66.98$ & $62.17$ \\
                Part CE & $68.49$ & $64.17$ \\
                Part BCE & $\mathbf{69.82}$ & $\mathbf{65.59}$ \\ \hline
            \end{tabular}
            \label{tab:ds_type}
    \end{minipage}
\end{minipage}
\end{center}

\subsection{Few-shot Learning}

We could also organize our PartImageNet in an another way by splitting non-overlapping classes into training, validation and testing set, thus it naturally becomes a few-shot learning and transfer learning benchmark. The new split especially designed for Few-shot Learning contains 109 classes in training set, 19 classes in validation set and 30 classes in testing set. Unlike tieredImageNet \cite{ren2018meta}, we do not try to avoid semantic overlap between training and testing sets. The details of the split will be presented in the supplementary materials. By converting PartImageNet into a few-shot benchmark, it offers the community a chance to validate and research on the effects of parts in this domain.

\subsubsection{Baseline}

We follow the conventional setting in few-shot classification to resize images to 84 * 84 pixels and adopt Conv4 and ResNet-12 as backbones with respect to different methods. We conduct experiments on PartImageNet using MAML \cite{finn2017model}, Prototypical Networks \cite{snell2017prototypical}, RFS \cite{tian2020rethinking}, Meta-Baseline \cite{chen2021meta}, COMPAS \cite{he2021compas} and DeepEMD \cite{zhang2020deepemd}. Among them, MAML \cite{finn2017model} and Prototypical Networks \cite{snell2017prototypical} are classic few-shot classification methods, RFS \cite{tian2020rethinking} and Meta-Baseline \cite{chen2021meta} are representative works for large-training-corpus methods and meta-training methods respectively while COMPAS \cite{he2021compas} and DeepEMD \cite{zhang2020deepemd} are recent works based on exploitation of parts or key regions to facilitate few-shot learning. 

Table \ref{tab:few_shot} summaries the results of selected methods on PartImageNet and miniImageNet. As can be observed, recent methods can obtain similar performances on PartImageNet as miniImageNet which shows that PartImageNet itself can serves as a good benchmark for evaluating few-shot algorithms. Besides, we also observe that though PartImageNet theoretically contains more classes with part structures, models \cite{he2021compas,zhang2020deepemd} that claim to exploit part information do not show obvious advantages compared to others when training without directly using the part annotations.

\begin{table*}
\small
\centering
\caption{Experimental results of Few-shot Learning on PartImageNet and miniImageNet. Average classification accuracies(\%) are reported.}
\label{tab:few_shot}
\begin{tabular}{cccccc}
\hline
\multirow{2}{*}{Model} & \multirow{2}{*}{Backbone} & \multicolumn{2}{c}{PartImageNet 5-way} & \multicolumn{2}{c}{miniImageNet 5-way} \\ \cline{3-6} 
 &  & 1-shot & 5-shot & 1-shot & 5-shot \\ \hline
MAML \cite{finn2017model} & Conv4 & $46.9\pm1.4$ & $58.1\pm0.7$ & $48.7\pm1.8$ & $63.1\pm0.9$ \\
Prototypical Networks \cite{snell2017prototypical} & Conv4 & $50.0\pm0.6$ & $65.4\pm0.5$ & $49.4\pm0.8$ & $68.2\pm0.7$ \\
RFS \cite{tian2020rethinking} & ResNet-12 & $66.8\pm0.9$ & $81.7\pm0.6$ & $64.8\pm0.6$ & $82.1\pm0.4$ \\
Meta-Baseline \cite{chen2021meta} & ResNet-12 & $68.0\pm0.3$ & $82.7\pm0.2$ & $63.2\pm0.2$ & $79.3\pm0.2$ \\
COMPAS \cite{he2021compas} & ResNet-12 & $67.1\pm0.5$ & $82.3\pm0.2$ & $65.7\pm0.5$ & $82.0\pm0.3$ \\
DeepEMD \cite{xu2020attentional} & ResNet-12 & $67.3\pm0.6$ & $82.7\pm0.4$ & $65.9\pm0.8$ & $82.4\pm0.5$ \\ \hline
\end{tabular}
\end{table*}
\vspace{-10mm}

\subsubsection{Can Parts help Few-shot Learning?}

Based on the initial results, we are interested in whether parts can actually help few-shot learning. We take two representative methods-COMPAS \cite{he2021compas} and DeepEMD \cite{zhang2020deepemd} here to validate it. 

COMPAS \cite{he2021compas} originally constructs a part dictionary $D$ of important regions by using K-Means on the feature representations of the backbone and further builds a map dictionary $S$ of the spatial activation distribution of these regions. To exploit the part annotation on COMPAS, we first convert the part segmentation annotations into bounding boxes followed by using pre-trained backbone to extract feature representations of these parts. Then we apply K-Means on the feature representations to obtain a better initialization of the part dictionary $D$. Similarly, we directly calculate the spatial distribution of these bounding boxes to offer a better initialization of the map dictionary $S$.

For DeepEMD \cite{zhang2020deepemd}, it originally tries to get a dense representations of images and then compute the Earth Mover's Distance to generate the optimal matching flows between the representation sets of images. The optimal matching cost is further used as the distance metric to measure the similarity of two images. In the updated version DeepEMD V2 \cite{zhang2020deepemdv2}, the authors find that instead of generating dense representations of images, it is better to randomly sample a set of regions in the images and only compute the EMD between these regions. Thus it is natural to replace these randomly generated regions with annotated part regions and their concatenated regions to see if the results get better.  

Table \ref{tab:few_part} summaries the results of COMPAS and DeepEMD w/ and w/o using part annotations on PartImageNet meta-testing set. Both methods achieve non-trivial improvement with explicit exploitation of annotated parts. Concretely speaking, COMPAS gets a $0.9\%$ and $0.6\%$ performance gain and DeepEMD achieves a $1.2\%$ and $0.9\%$ performance gain on 1-shot and 5-shot scenarios respectively. This reveals the great potential of introducing parts into few-shot learning. Potential research direction lies at how to integrate part detector into current few-shot learning pipeline. We leave such interesting works to the future.

\begin{table*}[ht]
    \small
    \centering
    \setlength{\abovecaptionskip}{0pt}   
    \setlength{\belowcaptionskip}{10pt}
    \caption{Experimental results of COMPAS and DeepEMD w \& w/o using part annotations. Average classification accuracies(\%) are reported.}
    \begin{tabular}{cccc}
    \hline
    \multirow{2}{*}{Model} & \multirow{2}{*}{Backbone} & \multicolumn{2}{c}{PartImageNet 5-way} \\ \cline{3-4} 
     &  & 1-shot & 5-shot \\ \hline
    COMPAS \cite{he2021compas} & ResNet-12 & $67.1\pm0.5$ & $82.3\pm0.2$ \\
    COMPAS w/ Part Annotations & ResNet-12 & $\mathbf{68.0\pm0.5}$ & $\mathbf{82.9\pm0.3}$ \\
    DeepEMD \cite{zhang2020deepemd} & ResNet-12 & $67.3\pm0.6$ & $82.7\pm0.4$ \\
    DeepEMD w/ Part Annotations & ResNet-12 & $\mathbf{68.5\pm0.7}$ & $\mathbf{83.6\pm0.3}$ \\ \hline
    \end{tabular}
    \label{tab:few_part}
\end{table*}
\vspace{-8mm}

\section{Conclusion}


Parts provide a good intermediate representation of objects that have many advantages. 
Once obtained, they can be exploited to increase the accuracy of recognition, localization and benefit many downstream tasks such as pose estimation. In this work, we introduce PartImageNet—a large, high-quality dataset with part annotation on a general set of classes. A set of new baselines are further set of different vision tasks including semantic part segmentation, object segmentation and few-shot learning. We further show that introducing parts is beneficial to object segmentation and few-shot learning. We also reveal that existing works have certain limitations which hinder them to produce satisfactory semantic part segmentation results under complex backgrounds and variations. We hope that with the propose of our PartImageNet, we could attract more attention to the research of part-based models to address these difficulties and make parts great again.
\\

\noindent \textbf{Acknowledgements.} The authors gratefully acknowledge
supports from NSF BCS-1827427 and ONR N00014-21-1-2812. AK acknowledges support via his Emmy Noether Research Group funded by the German Science Foundation (DFG) under Grant No. 468670075.

\clearpage
%
%
\bibliographystyle{splncs04}
\bibliography{egbib}
\end{document}